\documentclass[sigconf]{acmart}

\usepackage{booktabs} 
\usepackage{multirow}

\setcopyright{rightsretained}

\acmDOI{10.475/123_4}

\acmISBN{123-4567-24-567/08/06}

\acmConference[KDD]{}{2018}{London, United Kingdom}
\acmYear{2018}
\copyrightyear{2018}

\acmArticle{4}
\acmPrice{15.00}

\begin{document}
\title{Web-Scale Responsive Visual Search at Bing}

\author{Houdong Hu, Yan Wang, Linjun Yang, Pavel Komlev, Li Huang, \\ Xi (Stephen) Chen, Jiapei Huang, Ye Wu, Meenaz Merchant, Arun Sacheti}
\affiliation{%
  \institution{Microsoft}
  \streetaddress{One Microsoft Way}
  \city{Redmond}
  \state{Washington}
  \postcode{98052}
}
\email{{houhu, wanyan, linjuny, pkomlev, huangli, chnxi, jiaphuan, wuye, meemerc, aruns}@microsoft.com}

\renewcommand{\shortauthors}{H. Hu et al.}


\begin{abstract}
In this paper, we introduce a web-scale general visual search system deployed in Microsoft Bing.
The system accommodates tens of billions of images in the index, with thousands of features for each image, and can respond in less than $200$ ms.
In order to overcome the challenges in relevance, latency, and scalability in such large scale of data, we employ a cascaded learning-to-rank framework based on various latest deep learning visual features, and deploy in a distributed heterogeneous computing platform.
Quantitative and qualitative experiments show that our system is able to support various applications on Bing website and apps. 
\end{abstract}

%
%


\newcommand{\yan}[1]{\textcolor{blue}{{[Yan: #1]}}}
\newcommand{\houdong}[1]{\textcolor{green}{{[Houdong: #1]}}}
\newcommand{\kelly}[1]{\textcolor{red}{{[Kelly: #1]}}}
\newcommand{\stephen}[1]{\textcolor{magenta}{{[Stephen: #1]}}}

\maketitle

\section{Introduction}

\begin{figure*}[t]
\centering
\includegraphics[width=\textwidth]{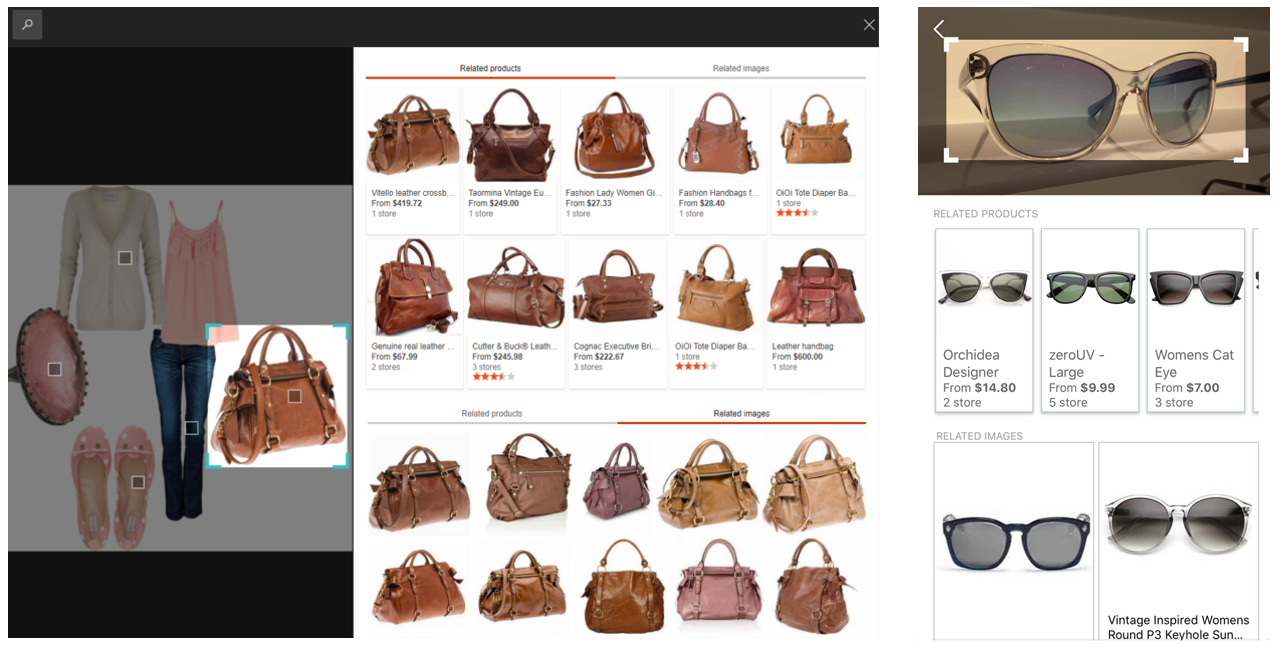}
\caption{Example user interfaces of the visual search system at Bing.
The left figure shows the desktop experience, where detected objects are shown as hotspots overlaid on the images.
Users are able to click a spot or specify their own crop box to get the visually similar products or images.
The right figure shows the mobile experience, where related products and images are shown for a query image of sunglasses.}
\label{fig:vsatbing}
\end{figure*}

\begin{figure*}[t]
\centering
\includegraphics[width=0.8\textwidth]{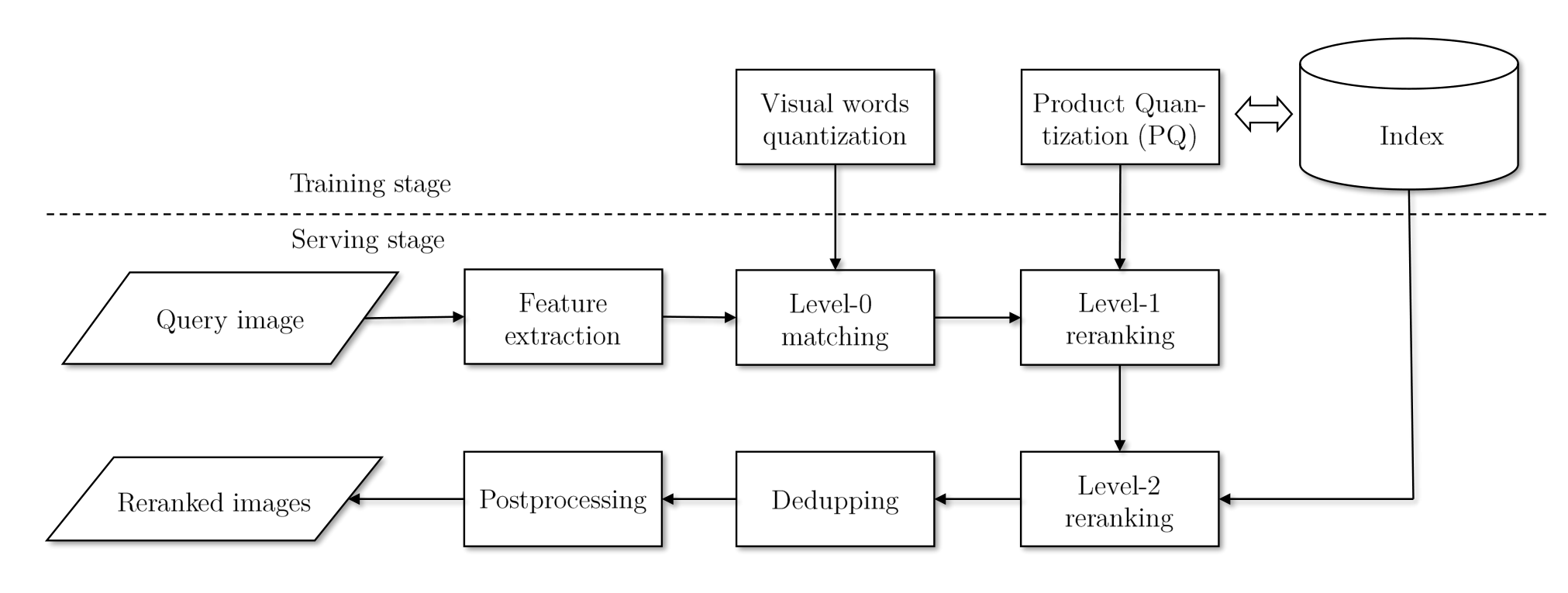}
\caption{Workflow overview of the web-scale visual search system in Bing.
The query image is first processed to be transformed into a feature vector, and then goes through a three-level cascaded ranker framework.
The result image list is returned to the user after postprocessing.
More details are available in Section~\ref{sec:SystemOverview}.}
\label{fig:overview}
\end{figure*}

Visual search, or Content-based Image Retrieval, is a popular and long-standing research area~\cite{EarlyYears,SpatialPyramidMatching,ThreeThings,VisualDiscoveryPinterest,VisualSearchEbay}.
Given an image, a visual search system returns a ranked list of visually similar images.
It associates a query image with all known information of the returned images, and thus can derive various applications, for example, locating where a photo was taken~\cite{im2gps}, and recognizing fashion items from a selfie~\cite{DeepFashion}.
Therefore, it is also of great interest in the industry.

Relevance is the main objective and metric of visual search.
With the recent development of deep learning~\cite{CBIRDeepLearningSurvey}, visual search systems have also got a boost on relevance, and have become more readily available for general consumers.
There has been exploration on visual search systems from industry players~\cite{VisualSearchEbay,VisualDiscoveryPinterest}, but the works focused more on the feasibility of vertical-specific systems, e.g. images on Pinterest or eBay, and lacked discussions on more advanced targets such as relevance, latency, and storage. 
In this paper, we would like to provide an overview of the visual search system in Microsoft Bing, hoping to provide insights on how to build a relevant, responsive, and scalable web-scale visual search engine.
To the best of the authors' knowledge, this is the first work introducing a general web-scale visual search engine.

Web-scale visual search engine is more than extending existing visual search approaches to a larger database.
Here web-scale means the database is not restricted to a certain vertical or a certain website, but from a spider of a general web search engine.
Usually the database contains tens of billions of images, if not more.
With this amount of data, the major challenges come from three aspects.
First, a lot of approaches that may work on a smaller dataset become impractical.
For example, Bag of Visual Words~\cite{VideoGoogle} generally requires the inverted index to be stored in memory for efficient retrieval.
Assuming each image has merely $100$ feature points, the inverted index will have a size of about $4$ TB, letting alone the challenges in doing effective clustering and quantization to produce reasonable visual words.
Second, even with proper sharding, storage scalability is still a problem.
Assuming we only use one single visual feature, the $4096$-dimension AlexNet~\cite{krizhevsky2012imagenet} \textit{fc7} feature, and shard the feature storage to $100$ machines, each machine still needs to store $1.6$ TB of features. 
Note these features cannot be stored in regular hard disks, otherwise the low random access performance will make the latency unacceptable.
Third, modern visual search engines usually use a learning-to-rank architecture~\cite{BoostRank,RankSVM} to utilize complimentary information from multiple features to obtain the best relevance. 
In a web-scale database, this posts another challenge on latency. 
Even the retrieval of the images can be parallelized, the query feature extraction, data flow control among sharded instances, and final data aggregation all require both sophisticated algorithms and engineering optimizations.
In addition to the aforementioned three difficulties from the index size, there is another unique challenge for general search engine.
The vertical-specific search engine usually has a controlled image database with well organized metadata.
However, this is not the case for general visual search engine, where the metadata is often unorganized, if not unavailable.
And this puts more emphasis on the capability to understand the content of an image.

In other words, it is very challenging to achieve high relevance, low latency, and high storage scalability at the same time in a web-scale visual search system.
We propose to solve the dilemma with smart engineering trade-offs.
Specifically, we propose to use a cascaded learning-to-rank framework to trade off relevance and latency, employ Product Quantization (PQ)~\cite{ge2013optimized} to trade off between relevance and storage, and use distributed Solid State Drive (SSD) equipped clusters to trade off between latency and storage scalability.
In the cascaded learning-to-rank framework, a sharded inverted index is first used to efficiently filter out ``hopeless'' database images, and generate a list of candidates.
And then more descriptive and also more expensive visual features are used to further rerank the candidates, the top of which is then passed to the final level ranker.
In the final level ranker, full-fledged visual features are retrieved and fed into a Lambda-mart ranking model~\cite{burges2010ranknet} to obtain similarity score between the query and each candidate image, based on which the ranked list is produced.

The remaining part of the paper is organized as follows.
In Section~\ref{sec:SystemOverview}, we will give a workflow overview of the entire system.
Then we will introduce the details together with engineering implementation of the system in Section~\ref{sec:Approach}, followed by applications introduced in Section~\ref{sec:Applications}.
Section~\ref{sec:Experiments} will provide quantitative and qualitative results of the proposed system in terms of relevance, latency and storage scalability, with conclusions in Section~\ref{sec:Conclusions}.

\begin{figure*}
\includegraphics[width=0.75\textwidth]{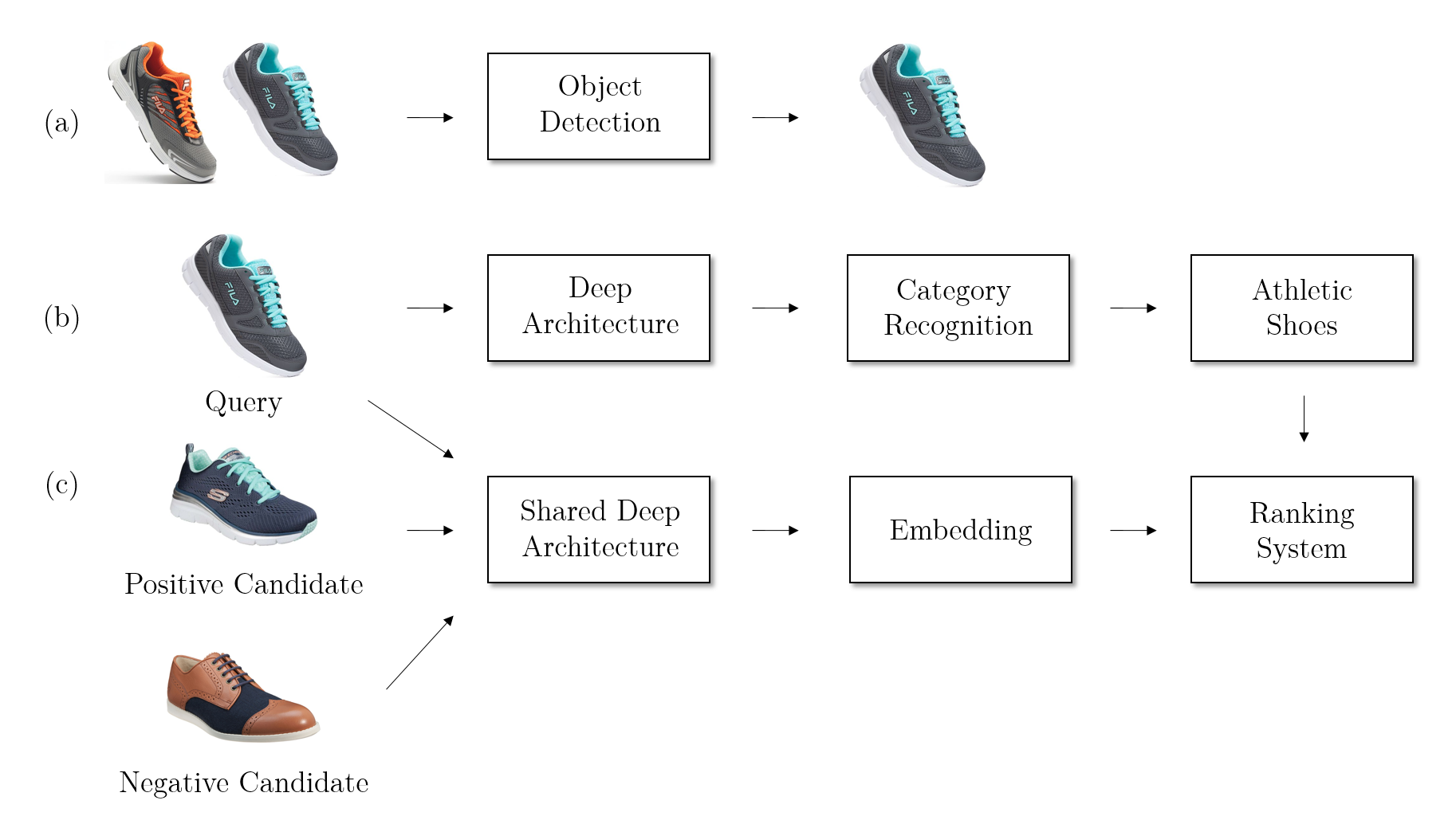}
\caption{The application scenarios of the DNN models used in the proposed system.}
\label{fig:DeepNeuralNetworkModels}
\end{figure*}

\section{System overview}
\label{sec:SystemOverview}

Before diving into how the system is built, let us first introduce the workflow.
When a user submits a query image that he/she finds on the Web or takes by a camera in Bing Visual Search,
visually similar images and products will be identified for the user to explore or purchase (examples are shown in Figure~\ref{fig:vsatbing}).
Bing Visual Search system comprises three major stages/components as summarized below, and Figure~\ref{fig:overview} illustrates the general processing workflow of how a query image turns to the final result images.

\textbf{Query understanding:}
We extract a variety of features from a query image to describe its content, including deep neural network (DNN) encoders, category recognition features, face recognition features, color features and duplicate detection features.
We also generate an image caption that can identify the key concept in the query image. 
Scenario triggering model is then called to determine whether to invoke different scenarios in visual search.
For instance, when a shopping intent is detected from the query, searches are directed to show a specific experience, rich in a particular segment of shopping.
Figure~\ref{fig:vsatbing} shows an example of the user interface for related product shopping scenario.
Bing Visual Search returns results with shopping information, such as store website links and prices.
Users can then follow the links to finish their purchase on the store websites. Object detection models will also run to detect objects, and users can click automatically tagged objects to view related products and related images.


\textbf{Image retrieval:}
The heavy lifting happens in the image retrieval module, which retrieves visually similar images based on the extracted features and intents.
As introduced before, the image retrieval process has a cascaded framework, dividing the entire process into Level-0 matching, Level-1 ranking and Level-2 ranking stages.
The details of the module will be introduced in Subsection~\ref{subsec:latency}.
After the ranked list is generated, we do postprocessing to remove duplicates and adult contents as needed.
This final result set is then returned to the user.

\textbf{Model training:}
Multiple models used in the retrieval process require a training stage.
First, several DNN models are leveraged in our system to improve the relevance. Each DNN model individually provides complementary information due to different training data, network structures and loss functions.
Second, a joint $k$-Means algorithm~\cite{xia2013joint} is utilized to build the inverted index in Level-0 matching. 
Third, PQ is employed to improve the DNN model serving latency without too much relevance compromise.
We also take advantage of object detection models to improve user experiences.
Details of these models are introduced in the following sections.

\section{Approach}
\label{sec:Approach}

In this section, we will cover the details of how we handle relevance, latency and storage scalability, including some extra features such as object detection.

\subsection{Relevance}

To optimize search relevance, we use multiple features, among which DNN features are most prominent.
We use several state-of-the-art DNNs including AlexNet~\cite{krizhevsky2012imagenet}, ZFSPPNet~\cite{zeiler2014visualizing, he2014spatial}, GoogleNet~\cite{GoogleNet, ioffe2015batch}, and ResNet~\cite{he2016deep}.
The last hidden layer is used as deep embedding features.
We train the networks using training data collected from multiple resources, including human labeling and web scraping, while it is also possible to leverage open source datasets. 
Our training datasets broadly cover thousands of major categories in daily life, and in-depth training datasets are also collected for specific domains, such as shopping.
We employ multiple loss functions to supervise the DNN training, such as softmax loss, pairwise loss and triplet loss. 
A complete set of the features we use is summarized in Table~\ref{tab:features in L2}.

Figure~\ref{fig:DeepNeuralNetworkModels} illustrates three examples of the DNN models we use.
Figure~\ref{fig:DeepNeuralNetworkModels}(a) is an object detection model, which does object localization and corresponding semantic category classification on the query image.
Figure~\ref{fig:DeepNeuralNetworkModels}(b) is a product classification network with a softmax loss.
We generate both the product category taxonomy and training data using segment specific repositories, which cover fashion, home furniture, electronics, and so on.
The category set includes thousands of fine-grained categories, such as $10$ subcategories in hats and caps including \emph{baseball caps}, \emph{sun hats}, and \emph{fedoras}.
We use Inception-BN network to obtain a trade-off between accuracy and latency.
The network is trained from scratch with various combinations of hyper learning parameters.
Both the \textit{pool5} layer DNN feature and the \textit{softmax} layer category feature are used in visual search ranker.
Figure~\ref{fig:DeepNeuralNetworkModels}(c) is a triplet network, which directly learns an embedding from images to an Euclidean space where distances between feature points correspond to image dissimilarity.
The network is trained on a large set of image triplets $(Q, I_1, I_2)$ annotated by human labelers, while each labeler answers questions about which image $I_1$, or $I_2$ is more similar to the query $Q$.
We fine-tune the triplet network on a pre-trained DNN network using ImageNet data, with the loss being minimized as
\begin{equation}
\begin{aligned}[rl]
L = \sum_{i=1}^{N} \Big( \lvert & f(x_{i}^\text{query}) - f(x_{i}^\text{pos}) \rvert^{2} \\
- & \lvert f(x_{i}^\text{query}) - f(x_{i}^\text{neg}) \rvert^{2} + \text{margin} \Big)_{+}.
\end{aligned}
\end{equation}
Here \(f(x)\) represents a compact embedding layer after last fully connected layer in original DNN, and this embedding layer is used as another DNN feature in visual search. 

Various DNN features, non-DNN visual features, object detection and text matching features are aggregated and leveraged in the visual search ranker.
More details are listed in Table~\ref{tab:features in L2}.
We also use local features ~\cite{winder2007learning} to remove duplicate images.
A Lambda-mart ranking model, which is a known multivariate regression tree model with a ranking loss function, takes the feature vector and produces a final ranking score.
The final result images are sorted by ranking scores, and are returned to the user.
The ranker training data collection process is similar as triplet training data collection.
An additional regression model is used to transform raw pair-wise judgments to list-wise judgments as the Lambda-mart ranking model training ground truth.

\begin{figure*}[t]
\centering
\includegraphics[width=0.9\textwidth]{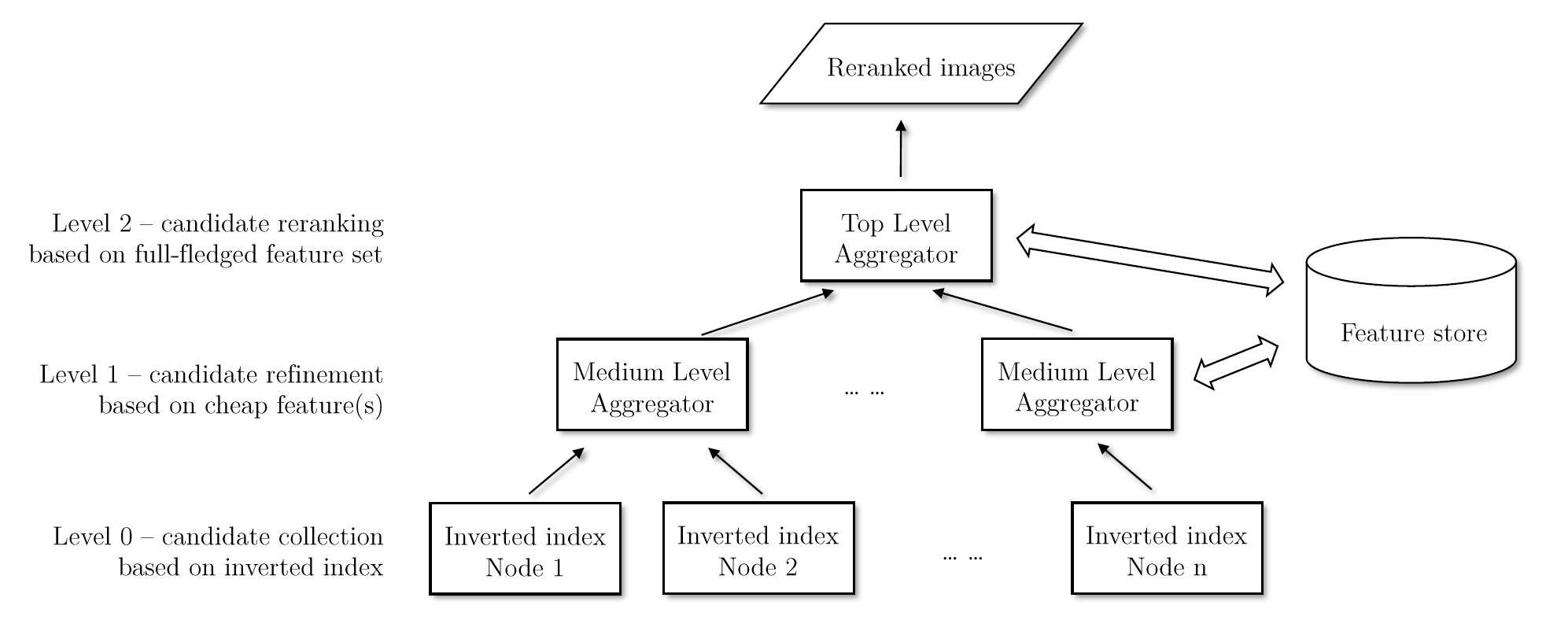}
\caption{Cascaded architecture of the web-scale visual search system in Bing.
Only the inverted index on Bag of Visual Words is used in Level-0 matching stage, to reduce the size of candidate set $10000$ times.
In Level-1 reranking stage, a single cheap DNN feature is used to further reduce the candidate number by $1000$ times.
Finally, full-fledged features are used to rerank the candidate images.}
\label{fig:cascaded}
\end{figure*}

\begin{table}
  \begin{tabular}{p{2.3cm}|p{5.0cm}}
    \hline
    Category & Feature  \\
    \hline
    \multirow{5}{*}{DNN} & AlexNet  \\
    & ZFSPPNet \\
    & GoogleNet \\
    & GoogleNetBN \\
    & ResNet-50 \\
    \hline
    \multirow{2}{*}{Visual} & Generic category  \\
    & Product category  \\
    & Face features \\
    & Dominant color \\
    \hline
    \multirow{2}{*}{Object detection} & Faster-RCNN \\
    & SSD \\
    \hline
    \multirow{2}{*}{Text matching} & Text matching with page metadata  \\
    & Text matching with click metadata  \\
    \hline
    \multirow{3}{*}{Deduplication} & Md5 \\
    & T2S2  \\
    & DupID \\
  \hline
\end{tabular}
\caption{Visual features used in Bing Visual Search.
Note the DNNs are trained with customized data and loss functions.}
\label{tab:features in L2}
\end{table}

\subsection{Latency}
\label{subsec:latency}

The image index contains tens of billions of images.
It is thus very challenging to provide responsive user experience.
A cascaded ranking framework is designed to achieve a good balance between relevance and latency.
The intuition is, filtering out ``hopeless'' candidates is much easier than retrieving visually similar images.
Therefore, cheap feature(s) can be used to significantly reduce the search space, followed by more expensive and expressive features  for high-quality reranking.
An inverted index can be used to generate the initial candidate set.

\begin{figure*}[t]
\centering
\includegraphics[width=0.7\textwidth]{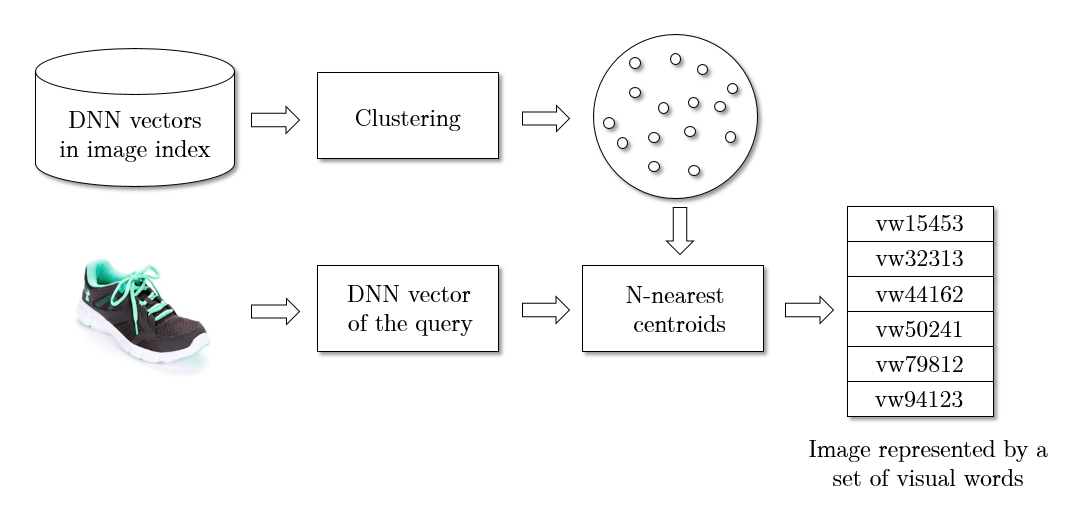}
\caption{Illustration of the training and matching stages of Bag of Visual Words.}
\label{fig:visualwords}
\end{figure*}

As shown in Figure~\ref{fig:cascaded}, the cascaded ranking framework includes Level-0 matching, Level-1 ranking and Level-2 ranking. 
Level-0 matching generates the initial candidate set based on an inverted index.
Following Arandjelovic et al.~\cite{ThreeThings}, we use visual words trained from joint $k$-Means as a cheap representation of an image, the detail of which is illustrated in Figure~\ref{fig:visualwords}.
After quantization, each image is represented by $N$ visual words.
Only index images with visual words matched with the query image are kept, and subsequently sent to the Level-1 ranker. 

\begin{figure*}[t]
\centering
\includegraphics[width=0.86\textwidth]{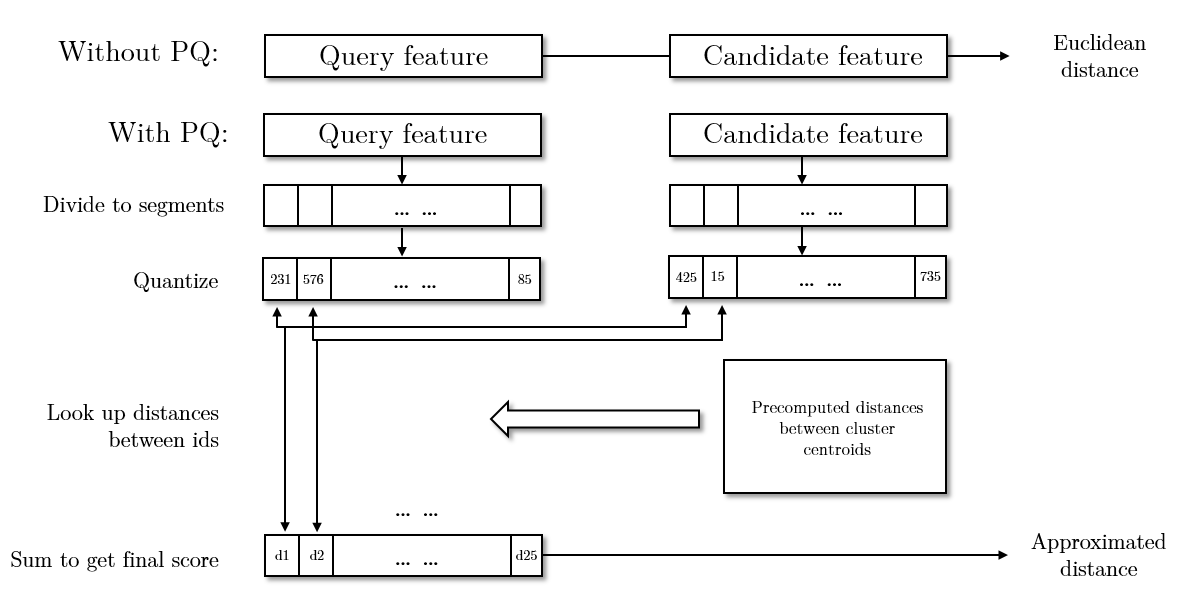}
\caption{Illustration of the calculation of PQ, with comparison to the approach without PQ.}
\label{fig:PQ}
\end{figure*}

The visual words narrow down a set of candidates from billions to several millions within several milliseconds.
However, it is still impractical to apply a full-fledged ranker discussed in the last section. 
Therefore, we use a light-wighted Level-1 ranker to further reduce the number of candidates.
A single DNN feature is calculated as a Level-1 ranker to further reduce the number of candidate index images.
Due to the cost of both storing the features and calculating the Euclidean distance between features, we employ PQ here.
The intuition is similar to Bag of Visual Words.
Instead of dealing with high-dimensional features, we use cluster centers to approximate the feature vectors.
The time cost of distance calculation can be dramatically reduced because the distance between cluster centers can be pre-computed.
The space cost can be saved as well because only IDs need to be stored, with a dictionary.
Figure~\ref{fig:PQ} provides an illustration of a high-dimensional vector decomposed into many low-dimensional sub-vectors to form a PQ vector.
Firstly we reduce the dimension of the DNN feature to $4n$ dimensions by Principal Component Analysis (PCA).
Then each new $4n$-dimensional DNN feature is divided into $n$ $4$-dimensional vectors, and the nearest cluster centroid from PQ model is determined for each $4$-dimensional vector.
Finally each DNN encoder can be represented by $n$ centroid IDs, which take only $n$ bytes, forming a PQ vector.
Then the feature distance can be approximately calculated as the distance between the query PQ vector $\{\text{PQ}^q_{i}\}_{i=1}^{n}$ and candidate image PQ vectors $\{\text{PQ}^c_{i}\}_{i=1}^{n}$. Note the Euclidean distance between each \(\text{PQ}^q_{i}\) and \(\text{PQ}^c_{i}\) could be pre-calculated and pre-cached.

As a result of Level-1 ranker, a candidate set can be reduced from millions to thousands.
At this point, a more expensive Level-2 ranker discussed in the last section is performed to rank the images more accurately.
All the DNN features used in Level-2 ranker also use PQ for acceleration. 

\subsection{Storage}

All the features used in our cascaded framework need to be stored in memory for efficient retrieval.
We represent DNN features as PQ IDs instead of real values, which greatly reduces storage requirement.
For example, the space requirement for a $2048$-dimensional ResNet DNN encoder reduces from $8$ KB space to $25$ bytes for each image by PCA and PQ, assuming $256$ clusters are used for each $4$-dimensional vector clustering and $n=25$ in PQ.
Visual words are used to reduce storage requirement in the matching step as well.
Specifically, one set of visual words occupies only $64$ bytes for each image.
Finally, we shard all the features, including the visual words inverted index, and PQ IDs into multiple (up to hundreds) machines to further save the calculation and storage burdens.
As also introduced in Subsection~\ref{subsec:latency}, we use a three-layer architecture to effectively manage the distributed index.

\subsection{Object Detection}

we utilize state-of-the-art object detection technology to provide easy and accurate guidance of visual search entry for users.
For instance, if the user is looking for outfit inspiration in a search engine, we predict the search/shopping intent of the user, and automatically detect several objects of user interests and mark them, so the user does not have to fiddle with the bounding box anymore.
We choose two approaches from several DNN based object detection frameworks for speed and accuracy balance: Faster R-CNN~\cite{ren2015faster} and Single Shot Multi-box Detection (SSD)~\cite{liu2016ssd}.
Various backbone structures and extensive parameter tuning have been experimented during model training to obtain the model with best precision and recall.
Fashion and home furniture are the first two segments that object detection models target on.
Object detection models will run only when specific visual intents are detected in the query image to save computational cost. 
In addition to providing accurate localization of the bounding box, the object detection module also gives inferred object category.
We use the detected category to match the existing image category in index to further improve the relevance.

\subsection{Engineering}

We profiled the entire visual search process, and found that the most time-consuming procedures were feature extraction and similar image retrieval (the Level-0, Level-1, Level-2 steps).
As a result, we leverage extensive engineering optimization in these two modules to reduce system latency and improve development agility.

\textbf{Feature extraction on CPU:}
While the project started when GPU acceleration was not common, a heavily optimized library called Neural Network Tool Kit (NNTK) was developed to optimize the DNN model deployment on CPU.
All core operations such as matrix multiplication and convolution are written in assembly language using latest vectorization instructions.
Convolution layer forward propagation is also carefully optimized for different channel sizes and feature map sizes to make them cache friendly.
we also use thread-level, machine-level parallelization and load balancing to further improve throughput.

Considering different application scenarios may demand different features, in order to allow flexibly reorganizing feature extraction pipelines without time-consuming recompilation, we develop a Domain Specific Language (DSL).
The DSL is executed with a graph execution engine to dynamically determine which operations need to be performed in order to obtain the asked features, and then execute in multiple threads.
For maximum reliability and flexibility, we implement the feature extraction system using Microsoft ServiceFabric, a micro-service framework, and deploy it on Microsoft Azure cloud service.
There is also a cache layer storing the features of all the images in the index, together with all the recently visited images by users.

\textbf{Feature extraction on GPU:}
With the emerge of production-quality GPU-accelerated DNN frameworks such as Microsoft Cognitive Toolkit (CNTK)~\cite{seide2016cntk} and Caffe~\cite{jia2014caffe}, we are one of the first teams deploying the serve stack of web-scale visual search systems on GPUs. 
GPU-based DNN evaluation is able to speed up the feature extraction by $10$ to $20$ times, depending on the model.
We deploy latency-sensitive models such as object detection to elastic GPU clusters on Azure, and achieve $40$ ms latency. 
Proper caching systems are able to further reduce the latency and cost.

\textbf{Efficient distributed retrieval:}
In addition to the cascaded framework, we also use a distributed framework to accelerate the image retrieval and reranking system, whose bottleneck is in Level-0 matching and feature retrieval.
The inverted index retrieval stage in Level-0 matching is performed in parallel on hundreds of machines, with aggressive timeout mechanism.
And the features are stored in SSDs which have significantly better random access performance than regular hard disks.

\section{Applications}
\label{sec:Applications}

Bing Visual Search has been deployed into both Bing.com and Bing App.
In both scenarios, the query image could be a web type image or a photo captured from the phone.
Bing Visual Search allows a user to specify an entire image as a search query by default.
It also supports a region of interest, so part of the image can act as the query.

\subsection{Bing.com}
In Bing.com, there are four entry points that would invoke visual search.

Suppose Alice is looking for kitchen decoration inspiration in Bing image search, and an image attracts her attention.
She clicks the image, and goes to the Image Details Page.
She can then find similar images in a related images section on the right side of Image Details Page (See the left column of Figure~\ref{fig:vsatbing}). 


Alice is particularly interested in a nice-looking chandelier in the image.
She can click the visual search button in the top right of the image, which is the second place to trigger visual search.
It will display a visual search box on the image.
Alice can drag and adjust the box to cover just the chandelier.
It will trigger related images search.
Bing also automatically detects users' search intent.
When a shopping intent is detected, in addition to regular image search, Bing also runs a product search to find matching products.
In this case, Bing will return a list of chandelier products.
Alice can simply click on the chandelier that is right for her, pick the best merchant in the related products section and finalize her purchase.
Every time she adjusts the visual search box, Bing automatically re-runs a visual search using the selected portion of the image as the query.

In the Image Details Page, Bing also detects objects automatically and marks them using overlaid hotspots.
So Alice does not have to adjust the bounding box manually -- just click the hotspot, and trigger a visual search.
This is considered as the third entry point.

In addition to the aforementioned three scenarios, users can also go to Bing image search page, and upload an image or paste an image URL by clicking the camera button inside the search bar.
Then visual search results can be displayed.


\subsection{Bing App}
We support visual search on mobile in two ways.
Firstly, we provide the same visual search functionalities on m.Bing.com as Bing.com. 
Secondly, users can install Bing's mobile app. 
Both m.Bing.com and Bing App support visual search using existing photos in the photo album, or directly taking a photo using the phone camera.
Besides it shows related products and related images, the essential goal is to excavate the knowledge behind the images using visual search techniques.

\section{Experiments}
\label{sec:Experiments}

In this section, we provide evaluations on the proposed system, which is deployed on Bing.
Following the major motivation, we use relevance, latency, and storage consumption to evaluate a general web-scale visual search engine.
We also provide some qualitative results to show the general result quality and experience of the proposed system.

\textbf{Relevance:} 
NDCG is used to evaluate the relevance.
A measurement set is prepared with thousands of queries sampled from Bing's search log.
With the queries, we obtain the most relevant candidate images we can get from Bing and other search engines.
Then we ask human labelers to do pairwise judgments on the relevance of the candidate images, based on which the ground truth ranking is calculated.
Based on this measurement set, we use the visual search system to rerank the candidate images, and compute NDCG@5 as the final metric. 

This metric has its advantages and drawbacks. 
An ideal way to evaluate the relevance of a search engine is to get a ground truth ranking between a certain query and all the images in the index.
With tens of billions images in the index, this is obviously infeasible.
That is where the proposed NDCG on a constrained measurement set has advantages: the approach makes the relevance evaluation feasible with acceptable labeling workload.
However, when analyzing the numbers, we also need to be aware that only the Level-2 reranking module is evaluated in this process, while index size, Level-0 matching and Level-1 reranking do not affect the NDCG number.

In Table~\ref{tab:DcgResults}, we report the NDCG@5 for the proposed system and different baselines with only a single feature.
All the DNN models listed in Table~\ref{tab:DcgResults} are trained with Bing dataset over thousands of generic categories.
From the table we can see that the chosen DNN features do provide complementary information and all contribute to improve the relevance.
We also analyze how the trade-off between storage and relevance affects relevance: the NDCG@5 of the system using raw DNN features, DNN features with dimensions reduced by PCA, and quantized DNN features by PQ on top of PCA are compared in the Table~\ref{tab:Resnet-50DcgResults}.

\textbf{Latency:}
We measure and report the End-to-End and component-wise system latency in Table~\ref{tab:latency}.
As introduced in Section~\ref{sec:Applications}, there are two different scenarios from a technical perspective.
One is when the query image is in our index, where visual discovery is the intended usage.
And the other is the query is uploaded by users, or when the user specifies a crop box to do visual search on a region of interest.
In the former case, because we already have the visual features, the system latency can be reduced by saving the network transferring and feature extraction time.
And the latter case will show a longer latency because we have to do full-fledged feature extraction.
As we can see from the table, because of the aggressive optimization on the cascaded reranking module, the latency bottleneck is actually the retrieval or calculation of visual features, depending on the scenario.
Note currently the feature extraction is still performed on CPU, which will be significantly accelerated if GPU is used.

\textbf{Storage:} We compare the storage cost per image with raw DNN feature, and quantized PQ ID in Table~\ref{tab:storage table}.
PQ generally shows $100$ to $1000$ times saving.

\begin{figure*}[t]
\centering
\includegraphics[width=1.0\textwidth]{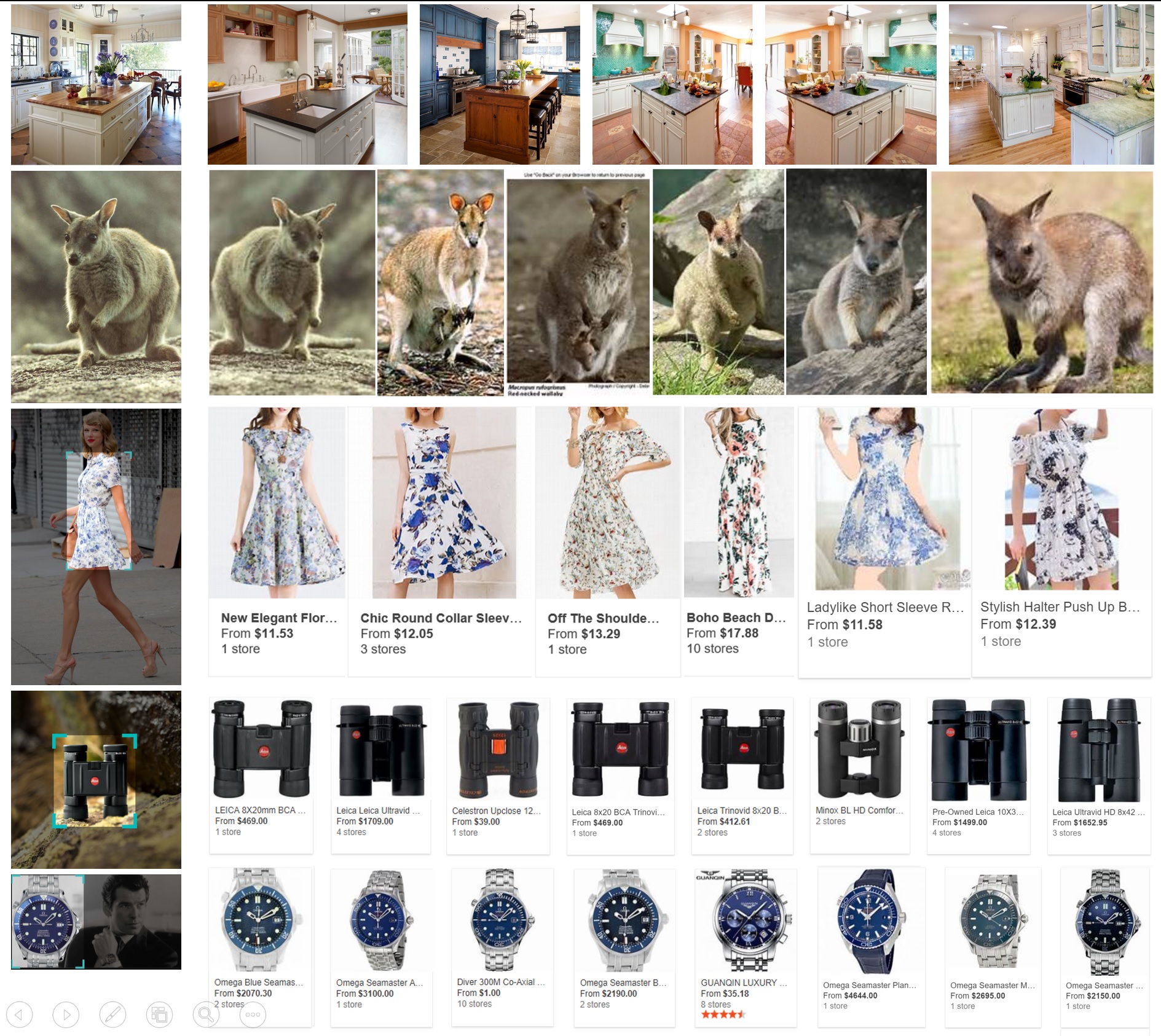}
\caption{Qualitative visual search results. The left column shows the query image, and the right column shows the returned results. The first two rows are from related images, and the other rows are from related products, with price and store information returned. } 
\label{fig:qualitativeresults}
\end{figure*}

Figure~\ref{fig:qualitativeresults} shows some examples of Bing Visual Search system, providing qualitative illustration of the system's performance on various scenarios, including queries with and without cropping boxes, related images and related products results.

\begin{table}
  \begin{tabular}{p{3.6cm}|p{2.5cm}}
    \hline
    Model & NDCG@5 \\
    \hline
    AlexNet \textit{fc7} + PQ & $58.87$ \\
    ZFSPPNet \textit{fc7} + PQ &  $64.68$ \\
    GoogleNetBN \textit{pool5} + PQ & $71.16$ \\
    ResNet-50 \textit{pool5} + PQ & $70.81$ \\ 
    \textbf{Proposed system} &  \bf{$74.20$} \\
  \hline
\end{tabular}
\caption{NDCG@5 on different systems.}
\label{tab:DcgResults}
\end{table}

\begin{table}
  \begin{tabular}{p{3.6cm}|p{2.5cm}}
    \hline
    Model & NDCG@5 \\
    \hline
    ResNet-50 \textit{pool5} & 71.84 \\
    ResNet-50 \textit{pool5} + PCA & 71.17  \\
    ResNet-50 \textit{pool5} + PQ & 70.81 \\ 
  \hline
\end{tabular}\caption{NDCG@5 on ResNet-50 raw feature, PCA and PQ.}
\label{tab:Resnet-50DcgResults}
\end{table}

\begin{table}
  \begin{tabular}{p{5.5cm}|p{1.5cm}}
    \hline
    Scenario/component & Latency \\
    \hline
    Visual Search (in-index image) & $174$ms \\
    Visual Search (user-uploaded image) & $1083$ms \\
    Object detection & $188$ms \\
    Feature extraction & $634$ms \\
    Cascaded ranking & $49$ms \\
  \hline
\end{tabular}
\caption{End-to-end and component-wise $50\%$ latency of Bing Visual Search on different baselines. The difference between in-index image scenario and user-uploaded image scenario is whether we can pre-compute the visual features and omit the network transfer.}
\label{tab:latency}
\end{table}

\begin{table}
  \begin{tabular}{p{2.5cm}|p{1.3cm}|p{0.5cm}|p{1.5cm}}
    \hline
    Model & Raw DNN Feature & PQ & Compression Ratio\\
    \hline
    AlexNet & 16384 & 25 & 655x \\
    ZFSPPNet & 16384 & 25 & 655x \\
    GoogleNet & 4096 & 25 & 164x \\
    GoogleNetBN & 4096 & 25 & 164x  \\
    ResNet-50 & 8192 & 25 & 328x \\ 
  \hline
\end{tabular}
\caption{DNN feature storage cost per image in bytes.}
\label{tab:storage table}
\vspace{-0.8cm}
\end{table}

\section{Conclusions}
\label{sec:Conclusions}

In this paper, we introduce approaches to overcome the challenges of a web-scale visual search system in relevance, latency, and storage scalability.
We use state-of-the-art deep learning features to achieve good relevance.
Training data in the form of pairwise judgment also contribute to improving relevance in the learning-to-rank framework.
We utilize both algorithms tailored for large-scale data and engineering optimization to achieve a latency less than $200$ ms and cost-effective storage.
Algorithms examples include PQ, inverted index, Lambda-mart ranker, and the cascaded ranker structure.
Engineering optimization includes GPU/CPU acceleration, distributed platform, and elastic cloud-based deployment.

There are many interesting directions one can pursue based on the proposed system.
The first is to further increase the index size.
The value of a search system as well as technical challenges increase dramatically when its scale improves.
Index containing more than hundreds of billion images may call for smarter encoding in features, more efficient index for feature retrieval, and more advanced models to achieve high relevance.
The second interesting direction is metric development. 
While relevance is a major motivation a user chooses a search engine over another, a set of diverse, attractive, and personal search results will also help user growth.
However, these expectations often go against DCG. 
How to define a comprehensive metric consistent with users' needs remains an important yet open question.
Last but not least, extending the system to leverage other devices such as FPGAs is another direction full of potential.
We learned many lessons when building Bing Visual Search, and see more challenges and opportunities along the way.

\begin{acks}

We thank Saurajit Mukherjee, Yokesh Kumar, Vishal Thakkar, Kuang-Huei Lee, Rui Xia, Tingting Wang, Aaron Zhang, Ming Zhong, Wenbin Zhu, Edward Cui, Andrey Rumyantsev, Tianjun Xiao, Viktor Burdeinyi, Andrei Darabanov, Kun Wu, Mikhail Panfilov, Vikas Cheruku, Surendra Ulabala, Bosco Chiu, Vladimir Vakhrin, Anil Akurathi, Lin Zhu, Aleksandr Livshits, Alexey Volkov, Mark Bolin, Souvick Sarkar, Avinash Vemuluru, Bin Wang, Alexandre Michelis and Yanfeng Sun for their supports.

\end{acks}

\bibliographystyle{ACM-Reference-Format}
\bibliography{sample-bibliography}

\end{document}